\documentclass[runningheads]{llncs}
\usepackage[T1]{fontenc}
\usepackage{graphicx}
\usepackage{amsmath}
\usepackage{amsfonts}
\usepackage{xcolor}
\usepackage{ulem}

\begin{document}
%
\thispagestyle{empty}
{\noindent\Large Springer Copyright Notice}\\[1pt]

{\noindent Copyright (c) 2023 Springer

\noindent This work is subject to copyright. All rights are reserved by the Publisher, whether the whole or part of the material is concerned, specifically the rights of translation, reprinting, reuse of illustrations, recitation,broadcasting, reproduction on microfilms or in
any other physical way, and transmission or information storage and retrieval, electronic adaptation, computer software, or by similar or dissimilar methodology now known or hereafter developed.}\\[1em]

{\noindent\large Accepted to be published in: 22nd International Conference on Image Analysis and Processing (ICIAP'23), Sept. 11--15, 2023.}\\\vfill

{\noindent Cite as:}\\[1pt]

{\setlength{\fboxrule}{1pt}
 \fbox{\parbox{0.95\textwidth}{S. F. dos Santos, R. Berriel, T. Oliveira-Santos, N. Sebe, and J. Almeida, ``Budget-Aware Pruning for Multi-Domain Learning,'' in \emph{22nd International Conference on Image Analysis and Processing (ICIAP'23)}, Udine, Italy, 2023, pp. 477--489, doi: 10.1007/978-3-031-43153-1\_40}}}\\\vfill

{\noindent BibTeX:}\\[1pt]

{\setlength{\fboxrule}{1pt}
 \fbox{\parbox{1.2\textwidth}{
 @InProceedings\{ICIAP\_2023\_Santos,

 \begin{tabular}{p{2mm}lcl}
  & author    & = & \{S. F. \{dos Santos\} and
               R. \{Berriel\} and
               T. \{Oliveira-Santos\} and\\ & & &
               N. \{Sebe\} and
               J. \{Almeida\}\},\\

  & title     & = & \{Budget-Aware Pruning for Multi-Domain Learning\},\\

  & pages     & = & \{477--489\},\\

  & booktitle & = & \{22nd International Conference on Image Analysis and Processing (ICIAP'23)\},\\

  & address   & = & \{Udine, Italy\},\\

  & month     & = & \{Sept. 11--15\},\\

  & year      & = & \{2023\},\\

  & publisher & = & \{\{Springer\}\},\\

  & doi       & = & \{10.1007/978-3-031-43153-1\_40\},\\
  \end{tabular}

\}
 }}}

\clearpage

\title{\scalebox{1.0}[1.0]{Budget-Aware Pruning for Multi-Domain Learning}\thanks{This research was supported by the FAPESP-Microsoft Research Virtual Institute (grant~2017/25908-6) and the Brazilian National Council for Scientific and Technological Development - CNPq (grant~314868/2020-8).
This work has been supported also by LNCC via resources of the SDumont supercomputer from the IDeepS project.}}

%
%

\author{Samuel Felipe dos Santos\inst{1}\orcidID{0000-0001-6061-5582} \and
Rodrigo Berriel\inst{2}\orcidID{0000-0002-6701-893X} \and
Thiago Oliveira-Santos\inst{2}\orcidID{0000-0001-7607-635X}
\and Nicu Sebe\inst{3}\orcidID{0000-0002-6597-7248}
\and Jurandy Almeida\inst{4}\orcidID{0000-0002-4998-6996}}

\authorrunning{Santos et al.}

%

\institute{Federal University of S\~{a}o Paulo, Brazil \\
\email{felipe.samuel@unifesp.br} \and
Federal University of Esp\'{i}rito Santo, Brazil \\
\email{berriel@lcad.inf.ufes.br, todsantos@inf.ufes.br} \and
University of Trento, Italy \\
\email{niculae.sebe@unitn.it} \and
Federal University of S\~{a}o Carlos, Brazil \\
\email{jurandy.almeida@ufscar.br} }

%
\maketitle              
\begin{abstract}

Deep learning has achieved state-of-the-art performance on several computer vision tasks and domains. 
Nevertheless, it still has a high computational cost and demands a significant amount of parameters.
Such requirements hinder the use in resource-limited environments and demand both software and hardware optimization. 
Another limitation is that deep models are usually specialized into a single domain or task, requiring them to learn and store new parameters for each new one. Multi-Domain Learning (MDL) attempts to solve this problem by learning a single model that is capable of performing well in multiple domains. Nevertheless, the models are usually larger than the baseline for a single domain.
This work tackles both of these problems: our objective is to prune models capable of handling multiple domains according to a user defined budget, making them more computationally affordable while keeping a similar classification performance.
We achieve this by encouraging all domains to use a similar subset of filters from the baseline model, up to the amount defined by the user's budget.
Then, filters that are not used by any domain are pruned from the network.
The proposed approach innovates by better adapting to resource-limited devices while, to our knowledge, being the only work that is capable of handling multiple domains at test time with fewer parameters and lower computational complexity than the baseline model for a single domain.

\end{abstract}
\section{Introduction}

Deep learning has brought astonishing advances to computer vision, being used in several application domains, such as 
medical imaging~\cite{zhou2022review}, autonomous driving~\cite{wang2021rod2021}, road surveillance~\cite{nguyen2020anomaly}, and many others.
However, to increase the performance of such methods, increasingly deeper architectures have been used~\cite{liu2021group}, leading to models with a high computational cost. Also, for each new domain (or task to be addressed), a new model is usually needed~\cite{berriel2019budget}. The significant amount of model parameters to be stored and the high GPU processing power required for using such models can prevent their deployment in computationally limited devices, like mobile phones and embedded devices~\cite{du2021bag}.
Therefore, specialized optimizations at both software and hardware levels are imperative for developing efficient and effective deep learning-based solutions~\cite{ISVLSI_2019_Marchisio}.

For these reasons, there has been a growing interest in the Multi-Domain Learning (MDL) problem.
The basis of this approach is the observation that, although the domains can be very different, it is still possible that they share a significant amount of low and mid-level visual patterns~\cite{rebuffi2017learning}.
Therefore, to tackle this problem, a common goal is to learn a single compact model that performs well in several domains while sharing the majority of the parameters among them with only a few domain-specific ones. This reduces the cost of having to store and learn a whole new model for each new domain.

Berriel~et~al.~\cite{berriel2019budget} point out that  one limitation of those methods is that, when handling multiple domains, their number of parameters is at best equal to the backbone model for a single domain. Therefore, they are not capable of adapting their amount of parameters to custom hardware constraints or user-defined budgets.
To address this issue, they proposed the modules named Budget-Aware Adapters (\textit{BA}$^2$) that were designed to be added to a pre-trained model to allow them to handle new domains and to limit the network complexity according to a user-defined budget.
They act as switches, selecting the convolutional channels that will be used in each domain.

However, as mentioned in~\cite{berriel2019budget}, although the use of this method reduces the number of parameters required for each domain, the entire model is still required at test time if it aims to handle all the domains. The main reason is that they share few parameters among the domains, which forces loading all potentially needed parameters for all the domains of interest. 

This work builds upon the \textit{BA}$^2$~\cite{berriel2019budget} by encouraging multiple domains to share convolutional filters, enabling us to prune weights not used by any of the domains at test time. Therefore, it is possible to create a single model with lower computational complexity and fewer parameters than the baseline model for a single domain. Such a model is capable of better fitting the budget of users with limited access to computational resources.

\begin{figure*}
    \centering
    \includegraphics[width=\textwidth]{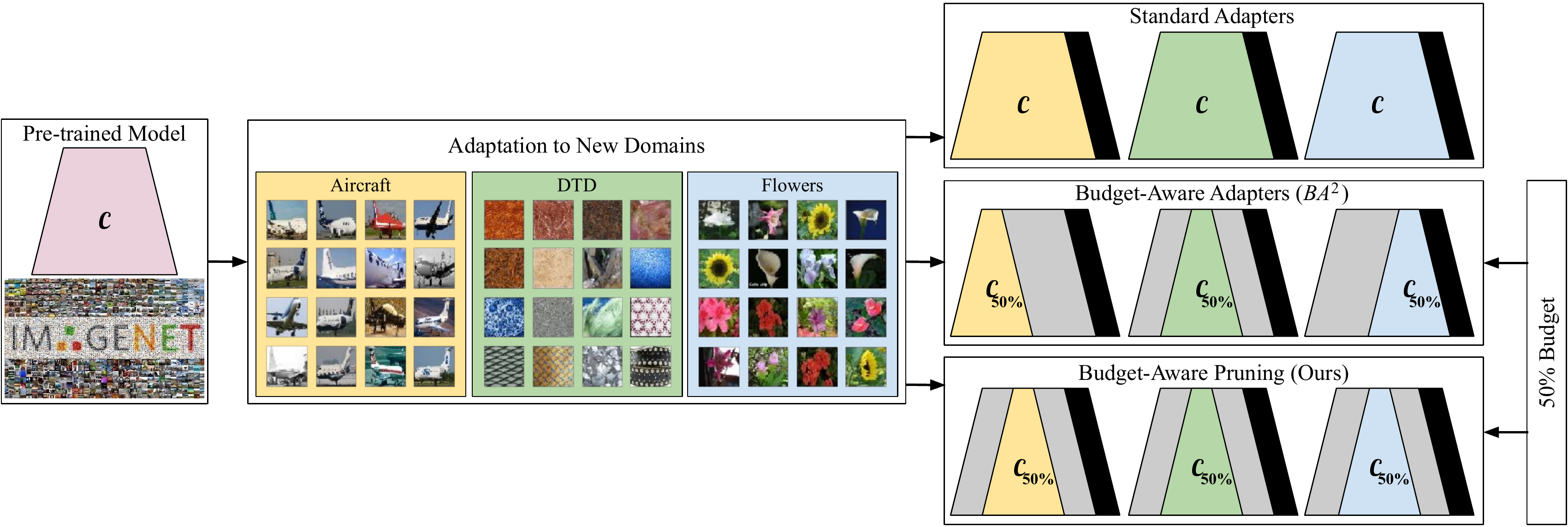}
    \caption{In standard adapters, the amount of parameters from the domain-specific models (indicated in colored $\mathcal{C}$) is equal to or greater than the backbone model (due to the mask represented in black).
    Budget-Aware Adapters can reduce the number of parameters required for each domain (unused parameters are denoted in gray). However, the whole model is needed at test time if handling distinct domains (colored areas share few parameters).
    Our model encourages different domains to use the same parameters (colored areas share most of the parameters). Thus, when handling multi-domains at test time, the unused parameters can be pruned without affecting the domains.}
    \label{fig:visao_geral}
\end{figure*}

Figure~\ref{fig:visao_geral} shows an overview of the problem addressed by our method, comparing it to previous MDL solutions and emphasizing their limitations.
As it can be seen, standard adapters use the entire model, while \textit{BA}$^2$~\cite{berriel2019budget} reduces the number of parameters used in each domain, but requiring a different set of parameters per domain. Therefore, the entire model is needed for handling all the domains together and nothing can be effectively pruned.
On the other hand, our approach increases the probability of using a similar set of parameters for all the domains.
In this way, the parameters that are not used for any of the domains can be pruned at test time.
These compact models have a lower number of parameters and computational complexity than the original backbone model, which facilitates their use in resource-limited environments.
To enable the generation of the compact models, we propose a novel loss function that encourages the sharing of convolutional features among distinct domains.
Our proposed approach was evaluated on two well-known benchmarks, the Visual Decathlon Challenge~\cite{rebuffi2017learning}, comprised of 10 different image domains, and the ImageNet-to-Sketch setting, with 6 diverse image domains.
Results show that our proposed loss function is essential to encourage parameter sharing among domains, since without direct encouragement, the sharing of parameters tends to be low.
In addition, results also show that our approach is comparable to the state-of-the-art methods in terms of classification accuracy, with the advantage of having considerably lower computational complexity and number of parameters than the backbone.

\section{Related Work}
\label{sec:related_work}

\begin{figure*}
    \centering
    \includegraphics[width=0.8\textwidth]{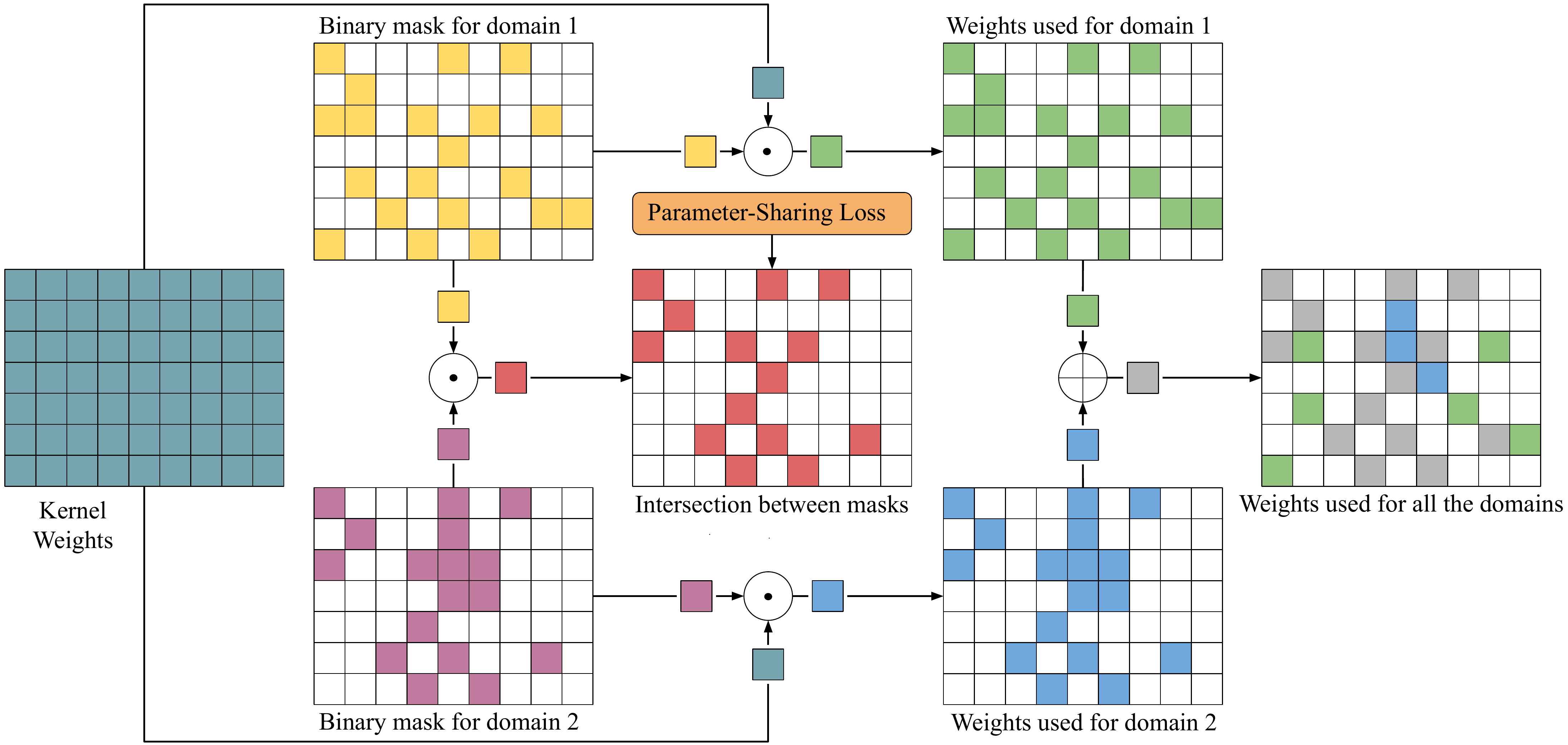}
    \caption{Overview of our strategy for sharing parameters among domains. 
    Colors represent data (i.e., weights, masks, etc), therefore, the colored squares denote the input data for each operation as well as its resulting output.
    }
    \label{fig:method_overview}
\end{figure*}

Previous approaches to adapt an existing model to a new domain used strategies like finetuning and pre-training, but faced the problem of catastrophic forgetting, in which the new domain is learned, but the old one is forgotten~\cite{hung2019increasingly}.
More recent MDL approaches usually leverage a pre-trained model as backbone. 
The backbone parameters are usually frozen and shared among all domains, while attempting to learn a limited and much lower amount of new domain-specific parameters~\cite{berriel2019budget}.
Approaches mostly differ from each other according to the manner the domain-specific parameters are designed, for example, domain-specific residual blocks and binary masks~\cite{berriel2019budget}.

For methods that use residual blocks to learn new domains, an example is the work of Rebuffi~et~al.~\cite{rebuffi2017learning,rebuffi2018efficient} that adds domain-specific parameters to the ResNet network in the form of serial or parallel residual adapter modules.

Following a different path, some works make use of binary masks to select different convolutional filters of the network for each domain, like the Piggyback method proposed by Mallya~et~al.~\cite{mallya2018piggyback}. 
In test time, the learned binary mask is multiplied by the weights of the convolutional layer.
Expanding on this idea, Mancini~et~al.~\cite{mancini2018adding,mancini2020boosting} also makes use of masks, however, 
they learn an affine transformation of the weights through the use of the mask and some extra parameters.  
Focusing on increasing the accuracy with masks, Chattopadhyay~et~al.~\cite{chattopadhyay2020learning} proposes a soft-overlap loss to encourage the masks to be domain-specific by minimizing the overlap between them. 

The works mentioned so far mainly focused on improving accuracy while attempting to add a small number of new parameters to the model, but they do not take into consideration the computational cost and memory consumption, making their utilization on resource-limited devices difficult~\cite{yang2022da3}.
Trying to address that, recent works have attempted to tackle the multi-domain learning problem while taking into account resource constraints.

Regarding parameters sharing, Wallingford~et~al.~\cite{wallingford2022task} proposed the Task Adaptive Parameter Sharing (TAPS), which learns to share layers of the network for multiple tasks by adding perturbations to the weights of the layer that are not shared. They also have a sparsity hyperparameter defined by the user.
Although this method lessen the amount of additional domain-specific parameters, it still always have considerably more parameters then the backbone model for a single domain.

Berriel~et~al.~\cite{berriel2019budget} proposed Budget-Aware Adapters (BA$^2$), which are added to a backbone model, enabling it to learn new domains while limiting the computational complexity according to the user budget.
The BA$^2$ modules are similar to the approach from Mallya~et~al.~\cite{mallya2018piggyback}, that is, masks are applied to the convolutional layers of the network, selecting a subset of filters to be used in each domain.
The network is encouraged to use a smaller amount of filters per convolution layer than a user-defined budget, being implemented as a constraint to the loss function that is optimized by constructing a generalized Lagrange function. 
Also, the parameters from batch normalization layers are domain-specific, since they perform poorly when shared.
This method and other continual learning strategies can reduce the number of parameters for a single domain.
However, these methods usually load the relevant parameters for the desired domain at test time. In order to load them for each domain of interest, it would be necessary to keep all the parameters stored in the device so that the desired ones are available. This way, the model does not fit the user needs, consuming more memory and taking more time to load, which might make it difficult to use in environments with limited computational resources. With this motivation we propose our method that encourages the sharing of parameters and is able
to effectively prune the model, reducing both the computational complexity and amount of parameters while handling all the domains.

\section{Pruning a Multi-Domain Model}
\label{sec:method}

This work was built upon the BA$^2$ modules from Berriel~et~al.~\cite{berriel2019budget} and proposes a new version to allow pruning unused weights at test time.
As results, the proposed method is able to obtain a pruned model that handles multiple domains, while having lower computational complexity and number of parameters than even the backbone model for a single domain. The pruned version is able to keep a similar classification performance while considering optimizations that are paramount for devices with limited resources.
Our user-defined budget allows the model to fit the available resources to a wider range of environments.
To achieve our goals, we added an extra loss function to BA$^2$ in order to encourage parameter sharing among domains and prune the weights that are not used by any domain. It was also necessary to train simultaneously in all the domains to be able to handle them all together at test time (see Figure~\ref{fig:method_overview} for an overview).

\subsection{Problem Formulation}
\label{sec:method_multi_domain}

The main goal of MDL is to learn a single model that can be used in different domains. One approach is to have a fixed pre-trained backbone model with frozen weights that are shared among all domains, while learning only a few new domain-specific parameters.
Equation~\ref{eq:backbone} describes this approach, where $\Psi_0$ is the pre-trained backbone model that when given input data $x_0$ from the domain $X_0$ return a class from domain $Y_0$ considering $\theta_0$ as the models weights.
Our goal is to have a model $\Psi_d$ for each domain $d$ that attributes classes from the domain $Y_d$ to inputs $x_d$ from the domain $X_d$ while keeping the $\theta_0$ weights from the backbone model and learning as few domain-specific parameters $\theta_d$ as possible.
\begin{align}
    \label{eq:backbone}
    \Psi_0 (x_0; \theta_0): X_0 \rightarrow Y_0\\
    \Psi_d (x_d; \theta_0, \theta_d ): X_d \rightarrow Y_d\notag 
\end{align}

Our starting point was the BA$^2$~\cite{berriel2019budget} modules, which are associated with the convolutions layers of the network, enabling them to reduce their complexity according to a user-defined budget.
Equation~\ref{eq:conv} describes one channel of the output feature map $m$ at the location $(i,j)$ of a convolutional layer, where $g$ is the activation function, $K \in \mathbb{R} ^ {(2K_H + 1) \times (2K_W + 1) \times C}$ is the kernel weights with height of $2K_H + 1$, width of $2K_W + 1$ and $C$ input channels, and $I \in \mathbb{R} ^ {H \times W \times C} $ is the input feature map with $H$ height, $W$ width and $C$ channels. 
\begin{align}
    \label{eq:conv}
    & m(i,j) = g( \sum^C_{c=1} \phi_c(i,j) )\\
    & \phi_c(i,j) = \sum^{K_h}_{h=-K_h} \sum^{K_w}_{w=-K_w} K(h, w, c) I(i-h, j-w, c)\notag 
\end{align}

Berriel~et~al.~\cite{berriel2019budget} proposed to add a domain-specific mask that is composed of $C$ switches $s_c$ for each input channel, as shown in Equation~\ref{eq:ba2}.
At training time, $s_c \in \mathbb{R}$ while, at test time, they are thresholded to be binary values. When $s_c = 0$, the weights $K_c$ (i.e., the filters for the $c$ input channel for a given output channel) can be removed from the computational graph, effectively reducing the computational complexity of the convolutional layers.
\begin{align}
    \label{eq:ba2}
    m(i,j) = g( \sum^C_{c=1} s_c \phi_c(i,j) )
\end{align}

The model is trained by minimizing the total loss $L_{total}$, which is composed of the cross entropy loss $L$ and a budget loss $L_B$, as shown in Equation~\ref{eq:ba2loss}, where $\beta \in [0,1]$ is a user-defined budget hyperparameter that limits the amount of weights on
each domain individually,  
$\theta_d^\beta$ are the domain-specific parameters for the budget $\beta$ and domain $d$, 
$\bar{\theta_d^\beta}$ is the mean value of the switches for all convolutional layers and  
$\lambda$ is the Karush-Kuhn-Tucker (KKT) multiplier.
\begin{align}
    \label{eq:ba2loss}
    L_{total} = L(\theta_0, \theta_d^\beta) + L_B( \theta_d^\beta, \beta)
\end{align}
The budget loss is given by $L_B( \theta_d^\beta, \beta) = \max ( 0, \lambda ( \bar{\theta_d^\beta} - \beta ) )$.
When the constraint $\bar{\theta_d^\beta} - \beta$ is respected, $\lambda=0$, otherwise, the optimizer increases the value of $\lambda$ to boost the impact of the budget.

\subsection{Sharing Parameters and Pruning Unused Ones}
\label{sec:method_sharing}

Although BA$^2$ can reduce the computational complexity of the model, it can not reduce the number of parameters necessary to handle all the domains together. Switches $s_c$ can only be pruned at test time when they are zero for \textit{all} domains, but they, in fact, assume different values if not forced to do so. 

For this reason, we added an additional parameter-sharing loss $L_{PS}$ to $L_{total}$, as described in Equation~\ref{eq:parameter_sharing}, where $N$ is the number of domains, $\theta_k^\beta$ for $k \in [1, ..., N]$ are the domain-specific parameters (switches) for each domain, $M$ is the total number of switches and $\lambda_{PS}$ is a hyperparameter that defines the importance of this loss component. 
\begin{align}
    \label{eq:parameter_sharing}
    L_{total} = L(\theta_0, \theta_d^\beta) + L_B( \theta_d^\beta, \beta) + L_{PS}( \theta_1^\beta, ..., \theta_N^\beta, \beta)  \\
    L_{PS}( \theta_1^\beta, ..., \theta_N^\beta, \beta) = \max(0, \lambda_{PS} ( 1-  \frac{|\theta_1^\beta \cap \theta_2^\beta \cap ... \cap \theta_D^\beta|)}{ M \beta}   ) \notag
\end{align}

The parameter-sharing loss calculates the intersection of all the domains masks and encourages it to grow up to the budget limitation.
Since the domain-specific weights from all the domains are required by this loss component, it is necessary to train on all of them simultaneously. Finally, the switches $s_c$ and the associated kernel weights $K_c$ can be pruned.

\section{Experiments and Results}
\label{sec:exp}

Our approach was validated on two well-known MDL benchmarks, the Visual Decathlon Challenge~\cite{rebuffi2017learning}, and the ImageNet-to-Sketch.

The Visual Decathlon Challenge comprises classification tasks on ten diverse well-known image datasets from different visual domains: 
ImageNet, Aircraft,  CIFAR-100, Daimler Pedestrian (DPed), Describable Textures (DTD), German Traffic Signs (GTSR), VGG-Flowers, Omniglot, SVHN, and UCF-101.
Such visual domains are very different from each other, ranging from people, objects, and plants to textural images.
The ImageNet-to-Sketch setting has been used in several prior works, being the union of six datasets: ImageNet, VGG-Flowers, Stanford
Cars, Caltech-UCSD Birds (CUBS), Sketches, and WikiArt~\cite{mallya2018piggyback}.
These domains are also very heterogeneous, having a wide range of different categories, from birds to cars, or art paintings to sketches~\cite{berriel2019budget}.

In order to evaluate the classification performance, we use the accuracy on each domain, and the S-score~\cite{rebuffi2017learning} metric.
Proposed by Rebuffi~et~al.~\cite{rebuffi2017learning}, the S-score metric rewards methods that have good performance over all the domains compared to a baseline, and it is given by Equation~\ref{eq:s_score}:
\begin{equation}
    \centering
    \label{eq:s_score}
    S = \sum_{d=1}^{10} \alpha_d \max\{ 0, Err_{d}^{max} - Err_{d} \}^{\gamma_d} 
\end{equation}
where $Err_{d}$ is the classification error obtained on the dataset $d$,  
$Err_{d}^{max}$ is the maximum allowed error from which points are no longer added to the score
and  $\gamma_d$ is a coefficient to ensure that the maximum possible $S$ score is $10.000$~\cite{rebuffi2017learning}.

To assess the computational cost of a model, we considered its amount of parameters and computational complexity.
For the number of parameters, we measured their memory usage, excluding the classifier and encoding float numbers in 32 bits and the mask switches in 1 bit.
For the computational complexity, we used the THOP library to calculate the amount of multiply-accumulate operations (MACs) for our approach, while we reported the values from \cite{berriel2019budget} for their work. All reported values are relative to the backbone size, as in \cite{berriel2019budget}.
Similar to \cite{berriel2019budget}, in order to assess the trade-off between effectiveness on the MDL problem and computational efficiency, we consider two variations of the S score, named as S$_O$, which is the S score per operation; and S$_P$, the S score per parameter.

We adopted the same experimental protocol of Berriel~et~al.~\cite{berriel2019budget}, making the necessary adjustments for our objective of pruning the model.

We used the SGD optimizer with momentum of 0.9 for the classifier and the Adam optimizer for the masks. 
All weights from the backbone are kept frozen, only training the domain-specific parameters (i.e., classifiers, masks, and batch normalization layers) and the masks switches were initialized with the value of 10$^{-3}$.
Data augmentation with random crop and horizontal mirroring with a probability of 50\% was used in the training phase, except for DTD, GTSR,  Omniglot, and SVHN, where mirroring did not improve results or was harmful.
For testing, we used 1 crop for datasets with images already cropped (Stanford Cars and CUBS), five crops (center and 4 corners) for the datasets without mirroring and 10 crops for the ones with mirroring (5 crops and their mirrors).
For the Visual Domain Decathlon, we used the Wide ResNet-28~\cite{zagoruyko2016wide} as backbone, training it for 60 epochs with batch size of 32, and learning rate of 10$^{-3}$ for the classifier and  10$^{-4}$ for the masks. Both learning rates are decreased by a factor of 10 on epoch 45.
For the ImageNet-to-Sketch setting, the ResNet-50 was used as backbone, training for a total of 45 epochs with batch size of 12, learning rate of 5$\times$10$^{-4}$ for the classifier and 5$\times$10$^{-5}$ for the masks, dividing the learning rates by 10 on epochs 15 and 30.

Differently from Berriel~et~al.~\cite{berriel2019budget}, we needed to train all the domains simultaneously, since we want to encourage the sharing of weights among them.
In order to do so, we run one epoch of each dataset in a round robin fashion. We repeat this process until the desired number of epochs are reached for each dataset. 

As ablation studies, we tested running BA$^2$ simultaneously on all tasks without the addition of our loss function, where we observed that there is a small drop in accuracy for doing so. This procedure is necessary since one must have information from all the domains at once to learn how to share parameters.
We also tested different strategies for simultaneous learning, for example, one batch of each domain, batches with data from multiple domains, among others. However, the effects on the results were small, so we chose the faster strategy, performing one epoch of each domain in a round-robin fashion with a random order.
We also performed a grid search on the validation set in order to select the best value for $\lambda_{PS}$, testing the values of 0.125, 0.25, 0.5, 0.75 and 1.0.  
For the Visual Domain Decathlon, the best $\lambda_{PS}$ was 1.0, while for the ImageNet-to-Sketch  it was $\lambda_{PS}=$0.125.

{After obtaining the best hyperparameter configuration, we compared our work to the baseline strategies of using the pre-trained model as a feature extractor, only training the classifier (named feature), and finetuning one model for each domain (finetune). 
We also compared to the state-of-the-art method BA$^2$, since it is one of the only works that take into consideration computational cost constraints.
The main focus of our work is the scenario where there is a budget set by the user, and other works except BA$^2$ do not take into consideration this restriction.
Despite the lack of attention that tackling multi-domain learning with budget restrictions has received, it is a promising topic that is paramount for the application of these methods in environments with limit computational power.

Experiments were run using V100 and GTX 1080 TI NVIDIA GPUs, Ubuntu 20.04 distribution, CUDA 11.6, and PyTorch 1.12.

After obtaining the best hyperparameter configuration, the model was trained on both training and validation sets and evaluated on the test set of the Visual Domain Decathlon. The comparison of the results with baseline strategies and a state-of-the-art method, BA$^2$, is shown in Table~\ref{tab:acc_test}.


\begin{table*}[!htb]
	\centering
	\caption{Computational complexity, number of parameters, accuracy per domain, S, S$_O$ and S$_P$ scores on the Visual Domain Decathlon.}
	\label{tab:acc_test}
    \resizebox{1\columnwidth}{!}{
    \begin{tabular}{l|cc|cccccccccc|ccc}
        \hline
        \hline
        
        Method & FLOP & Params & ImNet & Airc. & C100  & DPed & DTD & GTSR & Flwr. & Oglt. & SVHN & UCF & S-score & S$_O$ & S$_P$ \\
        
        \hline
        \hline
        \multicolumn{12}{l}{Baselines~\cite{rebuffi2017learning}:}\\
        Feature  & 1.000 & 1.00  & 59.7 & 23.3 & 63.1 & 80.3 & 45.4 & 68.2 & 73.7 & 58.8 & 43.5 & 26.8 &  544 &  544 & 544 \\ 
        Finetune & 1.000 & 10.0  & 59.9 & 60.3 & 82.1 & 92.8 & 55.5 & 97.5 & 81.4 & 87.7 & 96.6 & 51.2 & 2500 & 2500 & 250 \\
        \hline
        
        \multicolumn{12}{l}{BA$^2$~\cite{berriel2019budget}:}\\
        $\beta=1.00$ & 0.646 & 1.03  & 56.9 & 49.9 & 78.1 & 95.5 & 55.1 & 99.4 & 86.1 & 88.7 & 96.9 & 50.2 & 3199 & 4952 & 3106 \\
        $\beta=0.75$ & 0.612 & 1.03  & 56.9 & 47.0 & 78.4 & 95.3 & 55.0 & 99.2 & 85.6 & 88.8 & 96.8 & 48.7 & 3063 & 5005 & 2974 \\
        $\beta=0.50$ & 0.543 & 1.03  & 56.9 & 45.7 & 76.6 & 95.0 & 55.2 & 99.4 & 83.3 & 88.9 & 96.9 & 46.8 & 2999 & 5523 & 2912 \\
        $\beta=0.25$ & 0.325 & 1.03  & 56.9 & 42.2 & 71.0 & 93.4 & 52.4 & 99.1 & 82.0 & 88.5 & 96.9 & 43.9 & 2538 & 7809 & 2464 \\
        \hline
        
        \multicolumn{12}{l}{Ours: }\\
        $\beta=1.00$ & 0.837 & 1.03  & 56.9 & 37.3 & 80.2 & 95.1 & 57.9 & 98.6 & 84.6 & 83.8 & 96.0 & 45.8 & 2512 & 3001 & 2438 \\
        $\beta=0.75$ & 0.645 & 0.921 & 56.9 & 42.6 & 75.3 & 95.0 & 56.1 & 98.6 & 82.8 & 87.2 & 96.0 & 44.7 & 2444 & 3789 & 2654 \\
        $\beta=0.50$ & 0.447 & 0.783 & 56.9 & 42.1 & 73.7 & 96.8 & 51.3 & 98.7 & 81.4 & 87.1 & 96.1 & 45.4 & 2552 & 5709 & 3259 \\
        $\beta=0.25$ & 0.238 & 0.531 & 56.9 & 33.6 & 67.9 & 95.3 & 44.9 & 98.2 & 75.1 & 87.4 & 96.1 & 43.0 & 1942 & 8159 & 3657 \\
        \hline
        \hline

    \end{tabular}
    }
\end{table*}

Compared to the baseline strategies, our method was able to vastly outperform the feature extractor only, while achieving similar S-score to finetune for the budgets of $\beta=$1.0, 0.75 and 0.50, but with almost 10 times less parameters.

Compared to BA$^2$, we obtained similar accuracy in most domains, but faced some drops in accuracy in some domains compared to \cite{berriel2019budget}. 
We believe the main reason for this drop in accuracy is the simultaneous training procedure, as we observed similar drop when switching from individual to simultaneous training without the addition of our loss function, but we kept it since it is necessary to enable parameter sharing.
The domains with the biggest accuracy drops were the smaller datasets, like aircraft, DTD, VGG-Flowers, and UCF-101.
Other works, like Rebuffi~et~al.~\cite{rebuffi2017learning,rebuffi2018efficient} also mention subpar performance on these datasets, identifying the problem of overfitting.

The S-score also dropped up to 687 points for the same issues. The drop is harsher since the metric was designed to reward good performance across all datasets, and the small datasets we mentioned had a subpar performance.
Despite facing small drops in accuracy and S-score, our method offers a good trade-off between classification performance and computational cost.

When comparing computational complexity (FLOP on Table~\ref{tab:acc_test}), for the budgets of $\beta=$ 1 and 0.75, the original BA$^2$ had lower values, but for the harsher budgets of $\beta=$ 0.5 and 0.25, our methods obtained the lower complexity.
This happens due to the fact that the original BA$^2$ tends to discard more weights than the demanded when the budget is higher, while our methods tend to stay closer to the amount defined by the budget.
It also must be noted that all our methods obtained lower complexity than the value defined by the budget, showing that it is a great tool to adapt a backbone model to the resources available to the user.

By comparing the S$_O$ metric, we can observe that both methods have a good trade-off between computational complexity and S-score, as this metric greatly increases as the budget is reduced, showing that the reduction in computational complexity is considerably greater than the loss in S-score. 
As expected, our method had better S$_O$ for the harsher budgets of $\beta=$ 0.50 and 0.25 while BA$^2$ achieved superior results on the budgets of $\beta=$ 1.00 and 0.75.

The main advantage of our proposed method is the reduction on the number of parameters of the model, as it is, to our knowledge, one of the only methods that is capable of tackling the problem of multiple-domain learning, while also reducing the number of parameters in relation to the backbone model. Other methods can reduce the amount of parameters for a single domain, but since the parameters are not shared, to handle all of them during test time, the entire model must be kept.
As we can see (column Params of Table~\ref{tab:acc_test}), the original BA$^2$ had similar amount of parameters to the backbone model, being 3\% more for all budgets.
For the budget of $\beta=$ 1.00, we obtained the same result, while for the budget of $\beta=$ 0.75 we reduce the amount of parameters compared to the backbone model in 7.9\%, for  budget $\beta=$ 0.50, the reduction was of 22.7\% and for the for budget of $\beta=$ 0.25 there were 49.9\% less parameters.
This results shows that our method was successfully in encouraging the sharing of parameters among domains and that this approach can lead to considerable reductions on the amount of parameters of the network. 
The S$_P$ metric also show this results, as for the budgets of $\beta=$ 0.50 and 0.25 our method was able to outperform BA$^2$ by considerably reducing the amount of parameter.

Table~\ref{tab:acc_benchmark2} shows the results obtained on the test set of the ImageNet-to-Sketch setting. 
Compared to the baseline strategies, our method once again outperformed the use of the feature extractor only. Both our method and BA$^2$ obtained lower S-score than the finetune, showing that this benchmark is challenging.

\begin{table*}[!htb]
	\centering
	\caption{Computational complexity, number of parameters, accuracy per domain, S, S$_O$ and S$_P$ scores for the ImageNet-to-Sketch benchmark.}
	\label{tab:acc_benchmark2}
	\resizebox{1\columnwidth}{!}{
    \begin{tabular}{l|cc|cccccc|ccc}
        \hline
        \hline
        Method & FLOP & Params & ImNet & CUBS & Cars  & Flwr. & WikiArt & Sketches & S-score & S$_O$ & S$_P$ \\
        
        \hline
        \hline
        \multicolumn{8}{l}{Baselines:~\cite{mallya2018piggyback}:}\\ 
        Feature    & 1.000 & 1.00 & 76.2 & 70.7 & 52.8 & 86.0 & 55.6 & 50.9 &  533 &  533 & 533 \\
        Finetune  & 1.000 & 6.00 & 76.2 & 82.8 & 91.8 & 96.6 & 75.6 & 80.8 & 1500 & 1500 & 250 \\
        \hline        
        \multicolumn{8}{l}{BA$^2$~\cite{berriel2019budget}:}\\
        $\beta=1.00$  & 0.700 & 1.03 & 76.2 & 81.2 & 92.1 & 95.7 & 72.3 & 79.3 & 1265 & 1807 & 1228\\
        $\beta=0.75$  & 0.600 & 1.03 & 76.2 & 79.4 & 90.6 & 94.4 & 70.9 & 79.4 & 1006 & 1677 &  977\\
        $\beta=0.50$  & 0.559 & 1.03 & 76.2 & 79.3 & 90.8 & 94.9 & 70.6 & 78.3 & 1012 & 1810 &  983\\
        $\beta=0.25$  & 0.375 & 1.03 & 76.2 & 78.0 & 88.2 & 93.2 & 68.0 & 77.9 &  755 & 2013 &  733\\
        \hline

        \multicolumn{8}{l}{Ours:}\\
        $\beta=1.00$ & 0.777 & 1.09 & 76.2 & 79.1 & 82.2 & 92.4 & 70.4 & 77.2 &  726 &  934 &  666\\
        $\beta=0.75$ & 0.601 & 0.92 & 76.2 & 80.2 & 86.0 & 92.5 & 73.5 & 78.2 &  844 & 1404 &  917\\
        $\beta=0.50$ & 0.412 & 0.71 & 76.2 & 80.0 & 87.4 & 89.9 & 75.8 & 77.8 &  909 & 2206 & 1280\\
        $\beta=0.25$ & 0.222 & 0.49 & 76.2 & 75.5 & 83.9 & 88.6 & 72.5 & 77.3 &  689 & 3103 & 1406\\
        \hline
        
    \hline
    \end{tabular}}
\end{table*}

Comparing to the original BA$^2$, our model faced some drops on accuracy and S-score. Looking at the domain individually, we can see that the smaller datasets (Cars and Flwr.) were the ones with the greater drops in accuracy, a problem that also occurred on the Visual Domain Decathlon due to overfitting~\cite{rebuffi2017learning,rebuffi2018efficient}.  

In relation to the computational complexity, our models were better than BA$^2$ for the budgets of $\beta=0.25$ and $\beta=0.5$ and  slightly worse for $\beta=0.75$ and $\beta=1.0$. This is also reflected on the $S_O$ score, as we got better results for the same budgets.
Once again, our models obtained lower computational complexity than what was defined by the budget, showing that they fit the user needs.

The main advantage of our method is that it is capable of having a lower number of parameters than the backbone, even when handling multiple domains, something that BA$^2$ and most works in literature are not capable.
For the budget of $\beta=0.5$ and $\beta=0.25$, we obtained a considerable lower amount of parameters than the backbone model, reducing in 21.7\% and 46.9\%, respectively. 
This is reflected on the S$_P$ metric, where we were able to outperform BA$^2$ by a considerable margin in these budgets, showing that our model is more efficient.

\section{Conclusions}
\label{sec:conclusions}

In this paper, we addressed the multi-domain learning problem while taking into account a user-defined budget for computational resources, a scenario addressed by few works, but of vital importance for devices with limited computational power.
We propose to prune a single model for multiple domains, making it more compact and efficient. To do so, we
encourage the sharing of parameters among domains, allowing us to prune the weights that are not used in any of them, reducing both the computational complexity and the number of parameters to values lower than the original baseline for a single domain.
Performance-wise, our results were competitive with other state-of-the-art methods while offering good trade-offs between classification performance and computational cost according to the user's needs.
In future work, we intend to evaluate different strategies for encouraging parameter sharing, and test our method on different network models and benchmarks. 


%
%
%
%

\end{document}